# Zero-Shot Multi-Label Classification of Bangla Documents: Large Decoders Vs. Classic Encoders

**Souvika Sarkar[1], Md. Najib Hasan[1], Santu Karmaker[2]**

[1] Accessible AI Lab (A2I Lab), School of Computing, Wichita State University
[2] Bridge-AI Lab, Department of Computer Science, University of Central Florida
souvika.sarkar@wichita.edu, mxhasan39@shockers.wichita.edu, santu@ucf.edu

## Abstract

Bangla, a language spoken by over 300 million native speakers and ranked as the sixth most spoken language worldwide, presents unique challenges in natural language processing (NLP) due to its complex morphological characteristics and limited resources. While recent Large Decoder Based models (LLMs), such as GPT, LLaMA, and DeepSeek, have demonstrated excellent performance across many NLP tasks, their effectiveness in Bangla remains largely unexplored. In this paper, we establish the first benchmark comparing decoder-based LLMs with classic encoder-based models for Zero-Shot Multi-Label Classification (Zero-Shot-MLC) task in Bangla. Our evaluation of 32 state-of-the-art models reveals that, existing so-called powerful encoders and decoders still struggle to achieve high accuracy on the Bangla Zero-Shot-MLC task, suggesting a need for more research and resources for Bangla NLP.

## 1 Introduction

Bangla is the sixth most spoken language worldwide, with over 300 million native speakers, accounting for nearly 4% of the global population[1]. Despite its widespread use, Bangla remains a low-resource language in the domain of Natural Language Processing (NLP) (Joshi et al., 2020), with limited benchmark studies and scarce linguistic resources. The rise of classic encoders, such as LASER (Artetxe and Schwenk, 2019) and LaBSE (Feng et al., 2020), alongside large decoder-based models such as GPT (Brown et al., 2020), BLOOM (Scao et al., 2022), LLaMA (Touvron et al., 2023), and DeepSeek (DeepSeek-AI et al., 2025), has driven significant advancements in multilingual NLP. However, while these models have demonstrated exceptional performance in high-resource languages such as English, Chinese, and Spanish, their effectiveness in low-resource languages like Bangla remains largely unexplored. This gap raises a crucial question: How well do LLMs perform on Bangla-specific tasks compared to classic encoder-based models?

To address this, we present a benchmarking study evaluating 32 state-of-the-art models on the Zero-Shot Multi-Label Classification (Zero-Shot-MLC) task for Bangla, including three classical encoders—LaBSE (Feng et al., 2020), LASER (Artetxe and Schwenk, 2019), and BanglaTransformers (Foysal, 2020)—and 29 large decoder-based LLMs. Our findings reveal that while a few models achieve F1 scores around 60%, most struggle with reliable classification, highlighting significant gaps in multilingual adaptability. This study provides the first systematic decoder vs. encoder evaluation for Bangla Zero-Shot-MLC, emphasizing the need for further research and resources in Bangla NLP.

## 2 Background and Related Works

**Bangla NLP and Zero-shot classification**: In the domain of Bangla NLP, several tasks have been explored by researchers in the past, including Information Extraction (Rahman et al., 2008; Uddin et al., 2019; Sharif et al., 2016), Machine Translation (Hasan et al., 2019; Anwar et al., 2009; Ismail et al., 2014), Named Entity Recognition (Banik and Rahman, 2018; Chaudhuri and Bhattacharya, 2008), Question Answering (Islam et al., 2016; Sarker et al., 2019; Kowsher et al., 2019), Sentiment Analysis (Uddin et al., 2019; Hassan et al., 2016; Alam et al., 2017), Spam and Fake Detection (Islam et al., 2019; Hussain et al., 2020) etc. Recently, Kowsher et al. (2022) introduce Bangla-BERT, a monolingual BERT model for the Bangla language. Bhattacharjee et al. (2022) presented BanglaBERT – a BERT-based (Devlin et al., 2019) Bangla NLU model pre-trained on 27.5GB data.

---

[1] Source: *List of languages by the number of native speakers*, Wikipedia, https://en.wikipedia.org/wiki/List_of_languages_by_number_of_native_speakers (accessed on 23 May 2023)

Transformer-based models have been frequently employed in Bangla NLP research, such as in abusive comment detection (Aurpa et al., 2022b), text classification (Alam et al., 2020), question answering (Aurpa et al., 2022a; Adnan and Anwar, 2022), machine translation (Dhar et al., 2022), image caption generation (Palash et al., 2022), etc. Researchers have also introduced different datasets such as BanglaNLG (Bhattacharjee et al., 2023), a comprehensive benchmark for evaluating natural language generation models in Bangla, a sentiment analysis dataset SentNoB (Islam et al., 2021), a dataset for Bangla fake news detection (Hossain et al., 2020).

**Large Decoder (LLM) Based Generative Classification**: Recent research has extensively studied the potential of large language models like ChatGPT, Flan, BLOOM, etc., for a wide range of applications. For example, researchers have shown the utility of ChatGPT in healthcare education (Sallam, 2023), programming bug solving (Surameery and Shakor, 2023), and machine translation (Jiao et al., 2023).

**Classic Encoder Based Discriminative Classification**: Various topic modeling-based approaches have been explored for zero-shot classification for the English language (Karmaker Santu et al., 2016). Similarly, Li et al. (2018); Zha and Li (2019) worked towards a dataless text classification. Veeranna et al. (2016), adopted pre-trained word embedding for measuring semantic similarity between a label and documents. Further endeavor has been spent on zero-shot learning using semantic embeddings by (Hascoet et al., 2019; Zhang et al., 2019; Xie and Virtanen, 2021).

**Uniqueness of our Work:** As mentioned above, the performances of recent LLMs on Bangla documents are mostly unknown. To address this research gap, we examined multiple state-of-the-art sentence encoders and LLMs for a well-defined NLP task, i.e., zero-shot multi-label classification, with an exclusive focus on Bangla documents. Our study lays a foundation for future research endeavors in this under-explored yet important direction.

## 3 Problem Statement

We adopt the Definition-Wild 0SHOT-TC methodology proposed by Yin et al. (2019) and later explored by Sarkar et al. (2023, 2022) for the English language, as defined below.

**Definition 1.** *0SHOT-MLC: Given a collection of documents denoted as $D = \{d_1, d_2, ..., d_n\}$, a user represented by $U$ and a set of **user-defined** labels denoted as $L_U = \{l_1, l_2, ..., l_m\}$ provided in **real-time**, classify each document $d_i \in D$ with zero or more labels from $L_U$, without further fine-tuning.*

The 0SHOT-MLC employs embeddings of documents and labels, and the detailed methodology for embedding-based zero-shot MLC is provided in Appendix A.1.

## 4 Experimental Setup

### 4.1 Dataset

We created a new benchmark corpus, BanglaNewsNet, by crawling a large collection of publicly available online news from a renowned Bangladeshi news website, https://www.prothomalo.com/. Each article here is already labeled with one or more labels by human annotators. For example an article titled, "মেসিকে রিয়াল বেতিসে চান তাঁর আর্জেন্টিনা দলের সতীর্থ" (Messi's Argentina team-mate wants him in Real Betis) is associated with labels "ফুটবল" (Football) and "আর্জেন্টিনা ফুটবল দল" (Argentina Football Team). We scraped the article titles, article texts, and labels from the website and further cleaned the dataset by merging similar/repetitive labels into a single label. See Table 1 for an overview of the dataset.

| Dataset Name | # of Articles | Avg. article length | Labels retained | Labels/article |
|---|---|---|---|---|
| BanglaNewsNet | 7245 | ≈2517 words | 21 | 1.345 |

Table 1: An overview of the BanglaNewsNet dataset

### 4.2 Evaluation Metric

For evaluation, we report the Precision, Recall, and $F_1$ scores. Specifically, for each article, the model-inferred label(s) were compared against the list of "gold" label(s) to compute the true positive, false positive, and false negative statistics, which were then aggregated to compute the final Precision, Recall, and micro-averaged $F_1$ score.

### 4.3 Classic Encoders Vs. Large Decoders

For baselines, we implemented sentence similarity-based classification using LASER (Artetxe and Schwenk, 2019), LaBSE (Feng et al., 2020), and BanglaTransformer (Foysal, 2020). Additionally, we evaluated large decoder models from 13 families of instruction-tuned LLMs, as listed in Ta-ble 2. For commercial LLMs (OpenAI, Google, Anthropic), we used their APIs, while open-source models were accessed via the Hugging Face Transformers library. Since APIs for BLOOM (Scao et al., 2022), FLAN-UL2 (Wei et al., 2021), and GPTNeoX (Black et al., 2022) were unavailable, we utilized their embeddings to enable direct comparison with traditional encoders. Beyond embedding-based approaches, we also evaluated LLMs on the Bangla Zero-Shot-MLC task using a prompting-based approach. Unlike embedding similarity methods, which require access to model representations, prompting enables direct classification using model-generated responses, making it a practical alternative for black-box LLMs.

| LLM Family | Model |
|---|---|
| Google (Team et al., 2024) | gemini-1.5-pro (~200B) gemini-1.5-flash (8B) gemini-1.0-pro (~200B) gemma-2 (27B) gemma-2 (9B) gemma-2 (2B) flan-ul2 (20B) |
| OpenAI (Brown et al., 2020) | gpt-3.5-turbo-0613 (~20B) gpt-4o-mini-2024-07-18 (~20B) gpt-3 (175B) |
| EleutherAI (Black et al., 2022) | gpt-neox-20b (20B) |
| MetaAI Llama3 (Touvron et al., 2023) | Llama-3.1-8B (8B) Llama-3.2-3B (3B) meta-Llama-3.2-11B-Instruct (11B) Llama-3.3-70B-Instruct (70B) |
| BanglaLlama (Zehady et al., 2024) | BanglaLLama-3.1-8b (8B) BanglaLLama-3.2-3b (3B) BanglaLLama-3-11b (11B) |
| MistralAI (Jiang et al., 2024) | Mixtral-8x7B-Instruct-v0.1 (56B) |
| DeepSeek (DeepSeek-AI et al., 2025) | DeepSeek-R1-Distill-Llama-8B (8B) DeepSeek-R1-Distill-Qwen-14B (14B) DeepSeek-R1-Distill-Qwen-32B (32B) |
| AllenAI (OLMo et al., 2024) | OLMo-2-1124-7B-RM (7B) |
| Qwen (Yang et al., 2025) | Qwen1.5-72B (72B) Qwen2.5-7B-Instruct-1M (7B) |
| Gryphe (Gryphe) | MythoMax-L2-13b (13B) |
| UpStage (Kim et al., 2023) | SOLAR-10.7B-Instruct-v1.0 (10.7B) |
| Anthropic (Anthropic, 2024) | claude-3-haiku-20240307 (~20B) |
| BigScience (Scao et al., 2022) | bloom (176B) |

Table 2: The list of Large Decoder families and their variants evaluated in Bangla Zero-Shot-MLC.

## 5 Results

This section presents the experimental results on the BanglaNewsNet dataset, evaluating models based on F1 scores. Table 3 compares embedding-similarity methods using sentence encoders and three large decoder-based models. We utilized two label embedding approaches, such as *Label name + Keywords* and *Explicit-Mentions*, detailed in A.3. Based on the findings of Sarkar et al. (2023), these embedding methods previously yielded the best results, leading us to focus exclusively on them in this study. Among encoders, LaBSE performed best, achieving an F1 score of around 40%, while BanglaTransformer performed slightly lower. For large decoder-based models, Flan-UL2, BLOOM, and GPT-NeoX failed to achieve strong results in 0SHOT-MLC on Bangla articles. While some models exhibited high recall, their low precision made the results less meaningful. Notably, no LLM surpassed LaBSE in embedding-based 0SHOT-MLC, highlighting the limitations of current LLMs in handling its linguistic complexities.

### 5.1 LLM Performance Across Model Sizes

Table 4 presents the performance of LLMs categorized by parameter size. Below is a summary of key observations for each model category:

**Small Models (<8B).** OLMo-7B-Instruct (Groeneveld et al., 2024) and Qwen-2.5-7B-Instruct (Team, 2024b) outperform other models in this category, yet their F1 scores (0.381 and 0.351, respectively) remain low. Surprisingly, none of the LLMs in this group exceed the classic encoder LaBSE (F1: 0.400), highlighting their limitations in context of Bangla Zero-Shot-MLC.

**Mid-Sized Models (8B - 10B).** Gemini-1.5-Flash (8B) (Team et al., 2024) (F1: 0.571) outperforms larger models, surpassing Gemma-2-9B (Team, 2024a) (F1: 0.544) and DeepSeek-R1-Distill-8B (Guo et al., 2025) (F1: 0.470).

**Intermediate Models (10B - 15B).** MythoMax-L2 (13B) (Gryphe) underperforms significantly (F1: 0.187). Llama 3.2 (11B) (Touvron et al., 2023) and BanglaLlama 3.2 (11B) (Zehady et al., 2024) achieve close scores (0.513 and 0.481), showing minimal benefits from language-specific tuning.

**Larger Models (15B - 50B).** GPT-4o Mini ( 20B) (Brown et al., 2020) and Claude-3 Haiku ( 20B) (Anthropic, 2024) perform well (F1: 0.588, 0.540). Gemma-2 (27B) (Team, 2024a) leads this category (F1: 0.593). DeepSeek R1 Distill (32B) (Guo et al., 2025) also performs well (F1: 0.550), reinforcing the effectiveness of model distillation.

**Largest Models (>50B).** Gemini-1.5-Pro (Team et al., 2024) and Gemini-1.0-Pro (Team et al., 2023) achieve the highest F1 scores (0.616, 0.589). Despite their size, Mixtral-56B-Instruct (Jiang et al., 2024) (F1: 0.305), Qwen-1.5 (Yang et al.,

| | Classical Encoder Models | | | | | | Large Decoder Models | | | | | |
|---|---|---|---|---|---|---|---|---|---|---|---|---|
| | LASER | | LaBSE | | BanglaTransformer | | Flan-UL2 | | Bloom | | GPTNeoX | |
| Label Embedding -> | Label+KW | Expl.-Ment. | Label+KW | Expl.-Ment. | Label+KW | Expl.-Ment. | Label+KW | Expl.-Ment. | Label+KW | Expl.-Ment. | Label+KW | Expl.-Ment. |
| F1 Score | 0.267 | 0.305 | 0.354 | 0.404 | 0.334 | 0.384 | 0.234 | 0.241 | 0.329 | 0.341 | 0.345 | 0.357 |

Table 3: F1 score comparison of embedding based 0SHOT-MLC across classical encoders and large decoders.

| <8B Parameters | | 8B - 10B Parameters | | 10B - 15B Parameters | | 15B - 50B Parameters | | >50B Parameters | |
|---|---|---|---|---|---|---|---|---|---|
| Model | $F_1$ | Model | $F_1$ | Model | $F_1$ | Model | $F_1$ | Model | $F_1$ |
| Llama 3.2 (3B) | 0.320 | Llama 3.1 (8B) | 0.476 | Llama 3.2 (11B) | 0.513 | GPT 3.5 Turbo (~20B) | 0.470 | Llama 3.3 (70B) | 0.558 |
| BanglaLlama 3.2 (3B) | 0.323 | BanglaLlama 3.1 (8B) | 0.424 | BanglaLlama 3.2 (11B) | 0.481 | GPT 4o Mini (~20B) | 0.588 | Gemini 1.0 Pro (~200B) | 0.589 |
| Gemma 2 (2B) | 0.280 | Gemini 1.5 Flash (8B) | 0.571 | MythoMax L2 (13B) | 0.187 | Gemma 2 (27B) | 0.593 | Gemini 1.5 Pro (~200B) | 0.616 |
| OLMo 7B Instruct (7B) | 0.380 | Gemma 2 (9B) | 0.544 | SOLAR 10.7B Instruct (10.7B) | 0.452 | Claude 3 Haiku (~20B) | 0.540 | Mixtral 56B Instruct (56B) | 0.305 |
| Qwen 2.5 7B Instruct (7B) | 0.351 | DeepSeek R1 Distill (8B) | 0.470 | DeepSeek R1 Distill (14B) | 0.498 | DeepSeek R1 Distill (32B) | 0.550 | Qwen 1.5 (72B) | 0.429 |
| | | | | | | | | GPT 3.5 (175B) | 0.537 |

Table 4: F1 score comparison of prompting-based approaches across models with varying parameter sizes.

| Prompt Design |
|---|
| **System setup** |
| The AI assistant has been designed to understand and categorize user input by the given labels. When processing user input, the assistant must predict the labels from one of the following pre-defined options: 'চাকরিবাকরি' (Job market), 'করোনাভাইরাস' (Coronavirus), 'চলচ্চিত্র ও তারকা' (Movies and celebrities), 'স্বাস্থ্য' (Health), 'ব্যাংক' (Banking), 'অর্থনীতি' (Economy), 'শিক্ষা' (Education), 'প্রাকৃতিক দুর্যোগ' (Natural disasters), 'আইন ও আদালত' (Law and justice), 'কূটনীতি' (Politics), 'শিল্প ও বাণিজ্য' (Industry and commerce), 'ভ্রমণ' (Travel), 'নকশা' (Design), 'ফুটবল' (Football), 'খাবারদাবার' (Food and dining), 'দেশ ও রাজনীতি' (Country and politics), 'আন্তর্জাতিক' (International), 'দেশের খবর' (Country news), 'রাশিয়া ইউক্রেন সংঘাত' (Russia-Ukraine conflict), 'ক্রিকেট' (Cricket), 'নারী' (Women). It is essential to note that an article may have multiple labels. If the user input is not relevant with any labels, the assistant should print nothing, indicating that the input does not align with the available categories. The agent MUST response with the following json format: {"Labels": ["List of labels"]} |
| **User** Taking into account the given Bangla article {ব্যাংক এশিয়া ২০১৪ সালে ব্যাংকিং সেবার বাইরে থাকা বিপুল জনগোষ্ঠীকে ব্যাংকিং সেবায় আনতে এজেন্ট ব্যাংকিং সেবা চালু করে। বর্তমানে রাষ্ট্রমালিকানাধীন ও বেসরকারি মিলিয়ে ৩১টি ব্যাংক এ সেবা দিচ্ছে। বর্তমানে এজেন্ট ব্যাংকিং সেবা গ্রহণকারীর সংখ্যা দেড় কোটির বেশি। এর মধ্যে ব্যাংক এশিয়ার গ্রাহক ৫৫ লাখের বেশি। এসব গ্রাহকের ৯২ শতাংশই গ্রামীণ জনগোষ্ঠী। আবার ৬২ শতাংশ গ্রাহকই নারী। সারা দেশে ব্যাংক এশিয়ার ৫ হাজার ৪০০-এর বেশি এজেন্ট আউটলেট রয়েছে, যাদের মধ্যে নারী এজেন্ট ৫৪০ জন। ..... (In 2014, Bank Asia introduced agent banking services to bring banking services to a large population that was outside the reach of traditional banking. At present, a total of 31 banks, including both public and private, are providing these services. The number of users availing agent banking services has surpassed 140 million. Among them, Bank Asia has more than 5.5 million customers, of which 92% are from rural areas. Furthermore, 62% of the customers are women. Throughout the country, there are more than 5,400 agent outlets of Bank Asia, including 540 outlets managed by female agents.....)}, predict the category or labels of this article from the list of mentioned labels. |
| **Assistant** {"Labels": "ব্যাংক" (Banking), "শিল্প ও বাণিজ্য" (Industry and commerce)} |
| **Directive**: Taking into account the given Bangla article {article text}, predict the category or labels of this article from the list of mentioned {labels}. Please remember to only respond in the predefined JSON format without any additional information. |

Table 5: Prompt design details for the Zero-Shot-MLC task on BanglaNewsNet.

2025) (F1: 0.429), and GPT-3.5 (Brown et al., 2020) (F1: 0.417) significantly underperform.

> *Interesting Findings.*
> - **Top-performing models.** Gemini-1.5-Pro ( 200B, F1: 0.616) and Gemma-2 (27B, F1: 0.593) achieve the highest scores.
> - **Bigger isn't always better.** Gemini-1.5-Flash (8B) outperforms larger models like DeepSeek R1 Distill (32B), Mixtral (56B), and GPT-3.5 (175B).
> - **Encoders still hold value.** The classic encoder LaBSE (F1: .404) suprisingly outperforms all 8B-based models, reinforcing its effectiveness in zero-shot classification.
> - **LLMs struggle with low-resource languages.** Despite strong performance in English, most models fail to generalize well in Bangla, exposing a multilingual gap.

## 6 Discussion and Conclusion

Our study takes an important step toward evaluating state-of-the-art models for low-resource languages, with a focus on Bangla. We benchmark classic encoders and large decoder-based LLMs for Zero-Shot Multi-Label Classification (Zero-Shot-MLC), uncovering key limitations in both approaches. While large decoders perform well in high-resource languages, their effectiveness in Bangla remains inconsistent, with most failing to achieve reliable classification despite their scale. Classical encoders, though more stable, did not show significant improvement, indicating that neither approach is fully optimized for Bangla and exposing critical gaps in multilingual adaptation.

This work bridges a crucial research gap and also reinforces the need for tailored approaches suited to the complexities of morphologically rich low-resource languages like Bangla. By evaluating both encoders and decoders, our research contributes to ongoing efforts to enhance their effectiveness for regional languages, paving the way for future advancements in multilingual NLP.

# Limitations

As a limitation, it is worth mentioning that we had restricted access to several large language models (LLMs) such as LaMDA (137B), Jurassic-1 (178B), ERNIE 3.0 Titan (260B), Gopher (280B), GPT-4, Megatron Turing NLG (530B) and DeepSeek-R1 (671B), which limited their inclusion in our experiments. Additionally, while models like ChatGPT were available through an API, their usage was constrained by cost, which scales with model size, making large-scale evaluations financially challenging. Our computational resources were also a limiting factor, as we could only run open-source models up to 70 billion parameters, restricting our ability to test larger models natively. As a result, for models exceeding this threshold, we had to rely solely on API-based access, further increasing cost constraints. Furthermore, hardware limitations affected our ability to fine-tune models efficiently, limiting our exploration of task-specific optimizations. Another constraint was dataset diversity. Our experiments were conducted on a single dataset, and broader evaluations are necessary to determine whether our findings generalize across different domains.

## A Appendix

### A.1 *Encoder & LLM* Based 0-shotMLC

Here, we describe the steps used in our 0-shot-MLC approach.

1. Input Document: The end user provides the article text, custom-defined labels, and a set of keywords (optional).
2. Embedding Generation: The article text, labels, and keywords are transformed into rich embedding by leveraging the power of language models and encoding methods, capturing the essence of their content.
   - **Article Embedding**: We embed the entire article with sentence encoders and LLMs in a single shot.
   - **Label Embedding**: We adopted two different ways for target label embedding: 1) *Label + Keyword*- Label embedding using label name and keywords, 2) *Explicit-Mentions*- Label embedding using article-text which contains explicit mentions of label names.

   The details of these embeddings have been discussed in the Appendix A.2, A.3. While there are certainly other embedding methods possible, based on the findings of the paper (Sarkar et al., 2023), these embedding combinations worked best previously, and hence, we focused only on the above embedding type.
3. Threshold-based Label Assignment: Next, we quantify the cosine similarity between the article embedding and the label embeddings. Labels are assigned to the article based on a specified threshold, indicating the presence of a strong association. By experimenting with different threshold values (ranging from 0-1), a comprehensive analysis is conducted.
4. Zero-Shot multi-label classification: The outcome of this classifier is the prediction of relevant label(s) for the given article.

### A.2 "Entire Article" based embedding

Encode the entire article using sentence encoders or LLMs in a single shot, including articles that are long paragraphs and consist of more than one sentence.

### A.3 Label Embedding Approaches

We have used 2 different approaches for computing label embedding. The consecutive sections discuss about different procedures for generating label embedding.

#### A.3.1 "Label name + Keywords" based embedding

Encode both label name and keywords, then average all embeddings to generate the final label embedding.

### A.3.2 "Explicit-Mentions" based embedding

First, extract all the articles explicitly mentioning the label/phrase using algorithm 1 for all labels. Then, for each label, generate embeddings of all articles which are explicitly annotated/classified with that label, then average them to obtain the ultimate label embedding.

---

**Algorithm 1** Article Annotation using Explicit Mention

---
1: **Input:** Article text, Label names and Keywords
2: **Output:** Articles annotated with explicit Label
3: **for** each article text **do**
4:     check whether the label name or set (at least 3) of the informative keywords are present or not in the corresponding article text
5:     **if** present **then** annotate the article with the explicit label
6:     **end if**
7: **end for**

---

### A.4 Baseline Sentence Encoders

This section presents a bird's-eye view of the sentence encoders and large language models we have used for our experiments.

- **Language-Agnostic SEntence Representations (LASER)**: LASER (Artetxe and Schwenk, 2019) is a sentence encoding model that generates language-agnostic representations. It is capable of encoding sentences from multiple languages into fixed-length vectors, enabling cross-lingual tasks and multilingual applications.
- **Language-agnostic BERT Sentence Embedding (LaBSE)**: LaBSE (Feng et al., 2020) is a language-agnostic model based on the BERT architecture. It provides sentence embeddings that capture the semantic meaning of sentences across different languages. LaBSE allows for cross-lingual understanding and transfer-learning tasks.
- **Bangla sentence embedding transformer**: This Bangla sentence transformer (Foysal, 2020) is specifically designed for the Bangla language. It utilizes a transformer-based architecture to encode Bangla sentences into meaningful representations, enabling various NLP tasks in Bangla text analysis. It was trained on 2,50,000 Bangla sentences(wiki) by sentence transformer. This work is inspired by Sentence-BERT: Sentence Embeddings using Siamese BERT-Networks (Reimers and Gurevych, 2019) technique.

### A.5 Large Language Models

- **BLOOM**: Scao et al. (2022) introduce BLOOM, a massive language model with 176 billion parameters. BLOOM is trained on 46 natural languages and 13 programming languages and is the result of a collaborative effort involving hundreds of researchers. BLOOM is a causal language model trained to predict the next token in a sentence. This approach has been found effective in capturing reasoning abilities in large language models. BLOOM uses a Transformer architecture composed of an input embeddings layer, 70 Transformer blocks, and an output language-modeling layer. The sequential operation of predicting the next token involves passing the input tokens through each of the 70 BLOOM blocks. To prevent memory overflow, only one block is loaded into RAM at a time. The word embeddings and output language-modeling layer can be loaded on-demand from disk.
- **GPT-NeoX**: The GPT-NeoX-20B paper, authored by the Black et al. (2022), introduce an architecture similar to GPT-3 but with notable differences. They utilize rotary positional embeddings for token position encoding instead of learned embeddings and parallelize the attention and feed-forward layers, resulting in a 15% increase in throughput. Unlike GPT-3, GPT-NeoX-20B exclusively employs dense layers. The authors trained GPT-NeoX-20B using EleutherAI's custom codebase (GPT-NeoX) based on Megatron and DeepSpeed, implemented in PyTorch. To address computational limitations, the authors reused the hyperparameters from the GPT-3 paper. In their evaluation, the researchers compared GPT-NeoX-20B's performance to their previous model, GPT-J-6B, as well as Meta's FairSeq 13B and different sizes of GPT-3 on various NLP benchmarks, including LAMBADA, WinoGrande, HendrycksTest, and MATH dataset. While improvements were desired for NLP tasks, GPT-NeoX-20B exhib-

- ited exceptional performance in science and math tasks.

- **Google (Gemini & Flan)**: (Team et al., 2024) introduced Gemini-1.5 Pro and 1.0 Pro models containing powerful multimodal capabilities, allowing them to process text, images, audio, and video at the same time. This makes them highly effective for real-time AI tasks like reasoning and code generation. With an optimized transformer-decoder and sparse mixture-of-experts techniques, they enhance efficiency while handling complex inputs. Their long-context attention feature also helps them retain and process longer pieces of information more effectively. Meanwhile, the Gemma family (2B, 9B, and 27B) is built for flexible deployment, whether on-device or in the cloud, making it ideal for fast, low-latency applications. Flan-UL2, an improved version of Flan-T5, uses Mixture-of-Denoisers (MoD) pre-training to push the boundaries of NLP tasks like classification, reasoning, and question answering. With 20 billion parameters, it outperforms models like T5 and GPT-3, excelling in zero-shot learning and chain-of-thought reasoning while achieving top results on major NLP benchmarks.

- **OpenAI (GPT Series)**: According to (Brown et al., 2020) GPT-4o, GPT-3.5 Turbo, and GPT-3 represent major leaps in AI language models. GPT-4o is the most advanced, handling text, images, and audio at the same time with better speed and accuracy, making it great for real-time AI applications like chatbots and coding assistants. GPT-3.5 Turbo is designed for efficiency, balancing cost and performance, which makes it popular for business AI tools and content generation. GPT-3, with its 175 billion parameters, was a game-changer in AI, setting the stage for today's models with its strong language understanding and reinforcement learning for better alignment with human values.

- **MetaAI (Llama3)**: (Touvron et al., 2023) mentioned Meta's Llama3 models Llama-3.1-8B, Llama-3.2-3B, and Llama-3.3-70B are designed for efficient, cost-effective AI deployment. They use adaptive tokenization and transformer pruning to reduce computational demands while maintaining strong performance. The Llama-3.2-11B Vision-Instruct model expands Meta's work in multimodal AI, integrating visual and language reasoning for applications like computer vision, medical imaging, and smart assistants. By open-sourcing its models, Meta promotes collaborative AI development and innovation in decentralized AI systems. With real-time processing capabilities, Llama3 models stand out for their speed and energy efficiency, making them ideal for on-premise AI, embedded systems, and low-latency conversational tools.

- **BanglaLlama**: (Zehady et al., 2024) introduced BanglaLlama which is a groundbreaking initiative aimed at improving NLP for the Bangla language, addressing the lack of high-quality AI models for low-resource languages. With models ranging from 3B to 11B parameters, it is trained specifically on Bangla text to better capture linguistic nuances, cultural context, and accuracy. By using advanced tokenization and dataset augmentation, BanglaLlama excels in translation, content creation, and conversational AI. Its fine-tuned instruction-following capabilities help make AI more inclusive, ensuring non-English languages are better represented in global AI advancements. Beyond language processing, BanglaLlama also plays a crucial role in reducing biases, adapting to different Bangla dialects, and preserving indigenous languages in AI systems.

- **MistralAI**: (Jiang et al., 2024) represents Mixtral-8x7B as a groundbreaking Mixture of Experts (MoE) model, significantly reducing computational complexity while enhancing inference efficiency. Unlike conventional transformer architectures, Mixtral activates only a subset of its parameters per forward pass, reducing memory footprint and improving scalability. This innovative architecture allows highly efficient parallelized computation, making it a top contender for large-scale enterprise AI applications, real-time NLP solutions, and multilingual text generation. MistralAI's MoE-based LLMs are widely recognized for their superior speed-to-accuracy trade-offs, positioning them as one of the most energy-efficient large-scale models in the industry.

- **DeepSeek**: (DeepSeek-AI et al., 2025) men-

- tioned DeepSeek-R1 series, is designed for maximum efficiency, using advanced knowledge distillation to maintain strong performance with fewer parameters. With techniques like quantization and structured pruning, these models deliver fast inference speeds and low-latency processing, making them perfect for AI applications with limited resources. Ideal for real-time NLP, enterprise automation, and AI assistants, DeepSeek-R1 ensures quick responses while keeping computational demands low, making it a great choice for businesses and developers focused on efficiency.
- **AllenAI**: (OLMo et al., 2024) designed OLMo-2-7B with a strong focus on explainable AI (XAI) and interpretability. Unlike traditional black-box AI models, it incorporates features like attention transparency and explainability layers, ensuring clearer insights into how it processes information. Optimized for research in linguistics, AI ethics, and decision-making, OLMo is a top choice for academics, policymakers, and those building transparent AI systems that prioritize human understanding and trust.
- **Alibaba**: According to (Yang et al., 2025) Alibaba's Qwen series, including Qwen1.5-72B and Qwen2.5-7B, is built for enterprise applications, specializing in complex reasoning, structured NLP, and industry-specific adaptability. These models are fine-tuned for tasks like financial analysis, healthcare AI, and multilingual document processing, making them highly versatile for business use. With strong instruction tuning, the Qwen models excel at knowledge-intensive tasks, delivering high factual accuracy and outperforming many mainstream models in industry automation and specialized AI applications.
- **Gryphe**: (Gryphe) designed Gryphe's MythoMax-L2-13B for creative content generation, interactive storytelling, and narrative coherence. This model incorporates fine-tuned stylistic awareness and logical consistency, making it a preferred choice for conversational AI, virtual assistants, and AI-driven journalism.
- **UpStage**: (Kim et al., 2023) designed SOLAR-10.7B for flexibility, excelling in instruction tuning, few-shot learning, and adapting to various AI tasks. With a structured fine-tuning approach, it performs exceptionally well in areas like legal text interpretation, scientific research support, and summarizing complex documents with context and accuracy.
- **Anthropic**: (Anthropic, 2024) built Claude-3 Haiku with a strong focus on ethical AI, safety, and human-guided learning (RLHF). It excels at maintaining context, ensuring fairness, and delivering reliable real-time conversations, making it a great fit for critical applications in healthcare, governance, and compliance-focused AI.

### A.6 Precision-Recall Trade-Off in LLMs for Bangla

The performance analysis of large language models (LLMs) for zero-shot multi-label classification (MLC) in Bangla highlights several important aspects, particularly the trade-offs between precision and recall, which we have explored in the appendix due to space constraints.

#### A.6.1 Sentence-Encoder Models

Looking at the baseline sentence-encoder-based approaches (Table 6), we see that BanglaTransformer achieves the highest $F_1$ score (0.334), but with an evident imbalance between precision and recall. This pattern is also observed in LASER and LABSE, where recall is consistently higher than precision. While higher recall means the model retrieves more relevant labels, it also increases the number of false positives, which is a common issue in low-resource languages where high-quality training data is limited.

#### A.6.2 Large-Scale Generative Models

Shifting to large-scale generative models (Table 7), we observe a noticeable improvement in $F_1$ scores across the board, with GPT-NeoX (Black et al., 2022) ($F_1$: 0.357) outperforming BLOOM and FLAN-UL2 (Wei et al., 2021). However, GPT-NeoX's (Black et al., 2022) precision (0.241) remains significantly lower than its recall (0.675), reinforcing a major trend seen in generative models: they tend to be highly recall-biased, favoring coverage over accuracy. This can be attributed to their training approach, which optimizes for broad knowledge retrieval rather than precise classification. While this makes them effective for open-ended generation tasks, it poses a challenge for multi-label classification, where specificity is crucial. A recall-heavy approach may work well in some cases, such as medical document classification, where missing a critical label could be costly, but for general-purpose classification tasks, such a model could introduce significant noise. This reinforces the need for additional fine-tuning or hybrid methods to improve precision without sacrificing recall. Examining LLMs across different parameter sizes (Table 8), we see a clear scaling trend, where larger models generally achieve higher $F_1$ scores but with increasing recall at the expense of precision. Gemini 1.5 Pro ( 200B) (Team et al., 2024) achieves the highest $F_1$ score (0.616), with a recall of 0.918 and precision of just 0.463. This means that while it effectively captures relevant labels, it also introduces substantial noise in classification. A similar pattern is seen with GPT-4o Mini ( 20B) (Brown et al., 2020) and Claude-3 Haiku ( 20B) (Anthropic, 2024), with recall values of 0.889 and 0.810, respectively. These models demonstrate strong generalization and retrieval capabilities but lack the specificity required for accurate multi-label classification. However, Gemma 2 (27B) (Team et al., 2024) and DeepSeek R1 Distill (32B) (DeepSeek-AI et al., 2025) achieve more balanced trade-offs, with precision scores of 0.450 and 0.418 and recall scores of 0.833 and 0.804, respectively. This suggests that well-optimized architectures and distillation techniques can enable mid-sized models to match or even surpass larger models in classification tasks. Interestingly, GPT-3.5 (175B) (Brown et al., 2020) underperforms with an $F_1$ score of 0.537, reinforcing that parameter size alone does not guarantee superior classification accuracy. This aligns with previous findings in NLP research, where training data quality, fine-tuning strategies, and task-specific optimizations often play a more significant role than raw model size.

### A.6.3 Final Thoughts and Key Findings

From a statistical standpoint, the precision-recall trade-off observed across models is highly indicative of their underlying architectures and training methodologies. Sentence encoders, while effective in recall-driven tasks, fail to deliver precise classifications due to their limited exposure to label dependencies. Instruction-tuned LLMs, on the other hand, benefit from broader generalization but often lack the necessary specificity for multi-label classification, leading to recall-heavy biases.

Notably, distilled models like DeepSeek R1 Distill (32B) (DeepSeek-AI et al., 2025) demonstrate a more balanced performance, suggesting that parameter-efficient architectures can compete with, or even outperform, larger models.

| Topic+Keywords Based Label Embedding ||||||||| 
| LASER ||| LaBSE ||| BanglaTransformer |||
| Precision | Recall | $F_1$ | Precision | Recall | $F_1$ | Precision | Recall | $F_1$ |
| --- | --- | --- | --- | --- | --- | --- | --- | --- |
| 0.162 | 0.750 | 0.267 | 0.282 | 0.477 | **0.354** | 0.224 | 0.648 | 0.334 |
| Explicit-Mention Based Label Embedding ||||||||| 
| LASER ||| LaBSE ||| BanglaTransformer |||
| Precision | Recall | $F_1$ | Precision | Recall | $F_1$ | Precision | Recall | $F_1$ |
| 0.193 | 0.724 | 0.305 | 0.300 | 0.617 | **0.404** | 0.276 | 0.635 | 0.384 |

Table 6: Performance comparison of baseline sentence encoder-based approaches.

| Topic+Keywords Based Label Embedding ||||||||| 
| FLAN-UL2 ||| BLOOM ||| GPT-NeoX |||
| Precision | Recall | $F_1$ | Precision | Recall | $F_1$ | Precision | Recall | $F_1$ |
| --- | --- | --- | --- | --- | --- | --- | --- | --- |
| 0.135 | 0.890 | 0.234 | 0.231 | 0.574 | 0.329 | 0.235 | 0.634 | 0.345 |
| Explicit-Mention Based Label Embedding ||||||||| 
| FLAN-UL2 ||| BLOOM ||| GPT-NeoX |||
| Precision | Recall | $F_1$ | Precision | Recall | $F_1$ | Precision | Recall | $F_1$ |
| 0.144 | 0.742 | 0.241 | 0.232 | 0.642 | 0.341 | 0.241 | 0.675 | 0.357 |

Table 7: Performance comparison of different large language models.

| Performance Comparison of LLMs with Varying Parameter Sizes ||||| 
| Parameter Size | Model | Precision | Recall | $F_1$ |
| --- | --- | --- | --- | --- |
| **<8B** | Llama 3.2 (3B) | 0.203 | 0.710 | 0.320 |
| | BanglaLlama 3.2 (3B) | 0.219 | 0.595 | 0.323 |
| | Gemma 2 (2B) | 0.164 | 0.942 | 0.280 |
| | OLMo 7B Instruct (7B) | 0.365 | 0.410 | 0.380 |
| | Qwens 2.5 7B Instruct (7B) | 0.218 | 0.906 | 0.351 |
| **8B to 10B** | Llama 3.1 (8B) | 0.325 | 0.886 | 0.476 |
| | BanglaLlama 3.1 (8B) | 0.289 | 0.790 | 0.424 |
| | Gemini 1.5 Flash (8B) | 0.416 | 0.905 | 0.571 |
| | Gemma 2 (9B) | 0.388 | 0.913 | 0.544 |
| | DeepSeek R1 Distill (8B) | 0.343 | 0.782 | 0.470 |
| **10B to 15B** | Llama 3.2 (11B) | 0.359 | 0.895 | 0.513 |
| | BanglaLlama 3.2 (11B) | 0.336 | 0.849 | 0.481 |
| | MythoMax L2 (13B) | 0.128 | 0.347 | 0.187 |
| | SOLAR 10.7B Instruct (10.7B) | 0.342 | 0.650 | 0.452 |
| | DeepSeek R1 Distill (14B) | 0.350 | 0.350 | 0.861 |
| **15B to 50B** | GPT 3.5 Turbo (~20) | 0.350 | 0.741 | 0.470 |
| | GPT 4o Mini (~20B) | 0.439 | 0.889 | 0.588 |
| | Gemma 2 (27B) | 0.450 | 0.833 | 0.593 |
| | Claude 3 Haiku (~20B) | 0.402 | 0.810 | 0.540 |
| | DeepSeek R1 Distill (32B) | 0.418 | 0.804 | 0.550 |
| **50B>** | Llama 3.3 (70B) | 0.401 | 0.921 | 0.558 |
| | Gemini 1.0 Pro (~200B) | 0.796 | 0.468 | 0.589 |
| | Gemini 1.5 Pro (~200B) | 0.463 | 0.918 | 0.616 |
| | Mixtral 56B Instruct (56B) | 0.194 | 0.630 | 0.305 |
| | Qwens 1.5 (72B) | 0.353 | 0.540 | 0.429 |
| | GPT 3.5 (175B) | 0.515 | 0.573 | 0.537 |

Table 8: Performance comparison of prompting-based approaches across models with varying parameter sizes.